\documentclass[%
 aip,
 amsmath,amssymb,
 reprint,%
]{revtex4-1}

\usepackage{graphicx}
\usepackage{dcolumn}
\usepackage{bm}

\usepackage[utf8]{inputenc}
\usepackage[T1]{fontenc}
\usepackage{mathptmx}

\draft 
\newcommand\new{}
\begin{document}

\title{Accuracy of neural networks for the simulation of chaotic dynamics: precision of training data vs precision of the algorithm}

\author{S. Bompas}
\affiliation{%
Laboratoire Collisions, Agr\'egats, R\'eactivit\'e, IRSAMC, Universit\'e de Toulouse, CNRS, UPS, France
}
\affiliation{%
Laboratoire de Physique Th\'eorique, IRSAMC, Universit\'e de Toulouse, CNRS, UPS, France
}
\author{B. Georgeot} 
\affiliation{%
Laboratoire de Physique Th\'eorique, IRSAMC, Universit\'e de Toulouse, CNRS, UPS, France
}
\author{D. Gu\'ery-Odelin}
\affiliation{Laboratoire Collisions, Agr\'egats, R\'eactivit\'e, IRSAMC, Universit\'e de Toulouse, CNRS, UPS, France
}


\begin{abstract}
We explore the influence of precision of the data and the algorithm for the simulation of chaotic dynamics by neural networks techniques. For this purpose, we simulate the Lorenz system with different precisions using three different neural network techniques adapted to time series, namely reservoir computing (using ESN), LSTM and TCN, for both short and long time predictions, and assess their efficiency and accuracy. {\new Our results show that the ESN network is better at predicting accurately the dynamics of the system, and that in all cases the precision of the algorithm is more important than the precision of the training data for the accuracy of the predictions.} This result gives support to the idea that neural networks can perform time-series predictions in many practical applications for which data are necessarily of limited precision, in line with recent results. It also suggests that for a given set of data the reliability of the predictions can be significantly improved by using a network with higher precision than the one of the data. 
\end{abstract}


\maketitle 

 
  \section{introduction}

{\new Techniques of machine learning have been shown lately to be efficient in a huge variety of tasks, from playing the game of Go \cite{alphago} to speech recognition \cite{speech} or automatic translation \cite{transl}. In many cases, such breakthroughs correspond to complicated tasks with complex decision-making processes. However, it was highlighted recently that such tools can also be useful in tasks which are much more adapted to standard algorithms, such as simulation of physical systems. The simulation of chaotic dynamical systems has been known for a long time to be one of the most demanding, since the instability of the system makes small errors increase exponentially with time (see e.g. \cite{Lieberman,Ottbook,Gutzwiller}). Nevertheless, it was shown in \cite{Haas,Ottchaos,OttPRL,Ottattractor} that a certain type of machine learning algorithms called reservoir computing \cite{reservoir} was able to forecast the evolution of chaotic dynamical systems, even of high dimensionality (see also the recent collection of articles Ref.\cite{specialissue}). Remarkably enough, the simulation is made from the time series of the previous states of the system, without solving explicitly the equations defining the model, in a way similar to the model-free time series analysis approach developed earlier \cite{Kanz}. In parallel, it was also shown that other types of neural networks may be efficient as well in predicting the behaviour of such systems, i.e. LSTM networks \cite{Gers,LSTMchaos} or deep artificial neural networks \cite{3body}. Some comparisons of the different methods have already been performed on test models \cite{OttNN,Lorenzcomp}. 

Chaotic dynamical systems are inherently unstable, with positive Lyapunov exponents. This implies that any small error in the trajectory is exponentially amplified by the dynamics, making long time simulation of a specific trajectory practically impossible. In particular, small round-off errors due to the finite precision of the computation unavoidably get quickly amplified; for example, it was shown that it leads to an effective irreversibility of the system even if the equations are formally reversible \cite{dima1,dima2}: the simulated trajectory fails to retrace the original one after some time. Such short-time predictions of a specific trajectory are similar to weather forecasting in meteorology, where one wants the evolution from a specific initial state. However, it is known that such limitations do not preclude the long time simulation of chaotic systems to describe with reasonable accuracy the typical behaviour of trajectories of the system, enabling e.g. to get information about the structure of attractors in dissipative systems. This is similar to climate simulations, which can give information about future climates several years in the future, well beyond the limits of weather forecasting.

The results obtained so far have shown that the different machine learning techniques implemented for such problems can simulate with good accuracy both the short-time and long-time behaviour of chaotic dynamics.} However, it is important for future applications to assess the accuracy of these techniques in a precise way. In this paper, we explore the role of precision of the data used for the training of the network and of the algorithm itself on the accuracy of the simulation. We do so on a specific case of reservoir computing (Echo State Network-ESN) as well as on two other standard machine learning techniques used in this context, commonly called LSTM \cite{LSTM} and TCN \cite{TCN} techniques. We compare the accuracy of these methods to the explicit integration of the equations of motion, both for short time and long time predictions of a well known chaotic system originating from meteorology, the Lorenz system. Our results show that the precision of the algorithm is more important than the precision of the training data for the accuracy of the simulation. This has interesting consequences for applications, since the precision of the algorithm is by far easier to control than the one of the training data. We also discuss the training by considering trajectories of different size and by computing the time required to train the networks.

\section{Systems studied}

The Lorenz system was introduced in 1963 by Edward Lorenz \cite{Lorenz} as an extremely simplified model of meteorology. It corresponds to a set of three nonlinear coupled equations for the variables $x,y$ and $z$ as a function of time:

\begin{eqnarray}
\dot{x}&= \sigma (y-x)  \nonumber\\
\dot{y}&= x(\rho-z) \nonumber \\
\dot{z}& = x  y- \beta  z.
\label{Lorenzeq}
\end{eqnarray}
Throughout the paper we choose the standard set of parameters: $\sigma = 10, \rho = 28$ and $\beta = 8/3$.


This nonlinear and dissipative model displays chaotic features. In particular, trajectories converge to a low dimensional but complex structure referred to as a strange attractor, and in this specific case as the Lorenz attractor.  

\begin{figure}[ht]
  \begin{center}
    \includegraphics[width=0.45\textwidth]{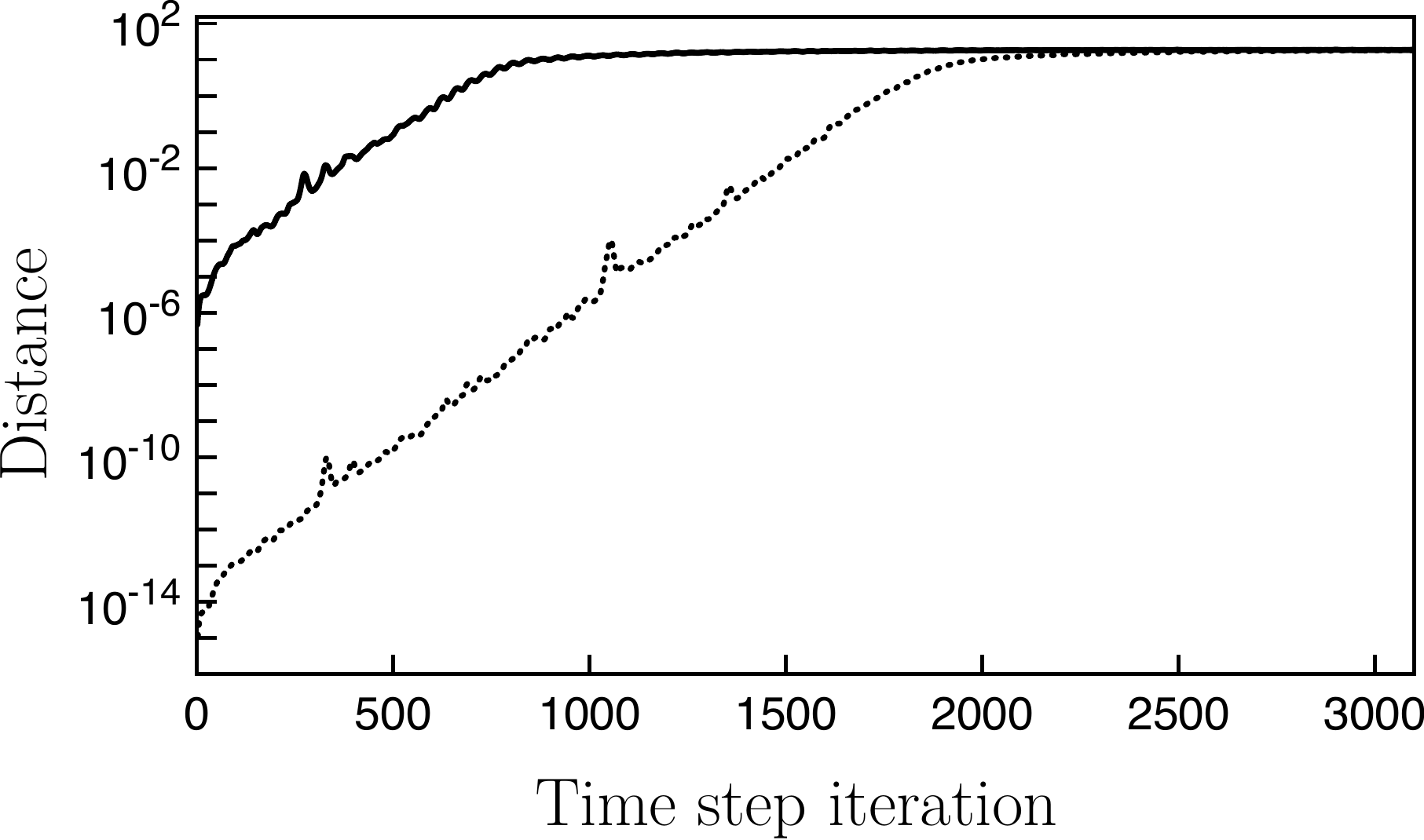}
  \end{center}  
  \caption{Euclidean distance between the reference trajectory of the Lorenz system obtained with quadruple precision with the double precision trajectory (dotted line) and the single precision trajectory (solid line), with a time step $dt=0.02$.
}
  \label{butterfly}
\end{figure} 

{\new As said in the introduction,} we distinguish two types of predictions. The short term predictions are similar to meteorological predictions: one starts from a specific initial point, and the aim is to follow a specific trajectory of the system for as long as possible. For strongly chaotic systems, this kind of predictions is limited by the exponential growth of perturbations: the distance between two nearby trajectories increases exponentially with time. This process, quantified by the (maximal) Lyapunov exponent, limits the numerical prediction of such systems since small imprecisions in the initial state will quickly increase to a macroscopic size. This phenomenon, noticed by Lorenz in the first paper on the system and often dubbed the ``butterfly effect''  is associated to a Lyapunov time which is logarithmic in the precision and sets a limit to numerical simulations with a given precision. This is shown in  Fig.~\ref{butterfly} in which we represent the distance to a reference trajectory  computed with a Runge-Kutta integration method of order 4 (RK4) in quadruple precision for trajectories computed using RK4 with lower precision (i. e. separated initially by $10^{-16}$ or $10^{-8}$). They strongly depart after a certain time from the high precision trajectory. The separation time clearly increases only logarithmically with the precision. 

This property makes numerical simulation of specific trajectories for chaotic systems very difficult: increasing by exponentially large factors the precision only increases linearly the prediction time.

\begin{figure}[ht]
  \begin{center}
    \includegraphics[width=0.45\textwidth]{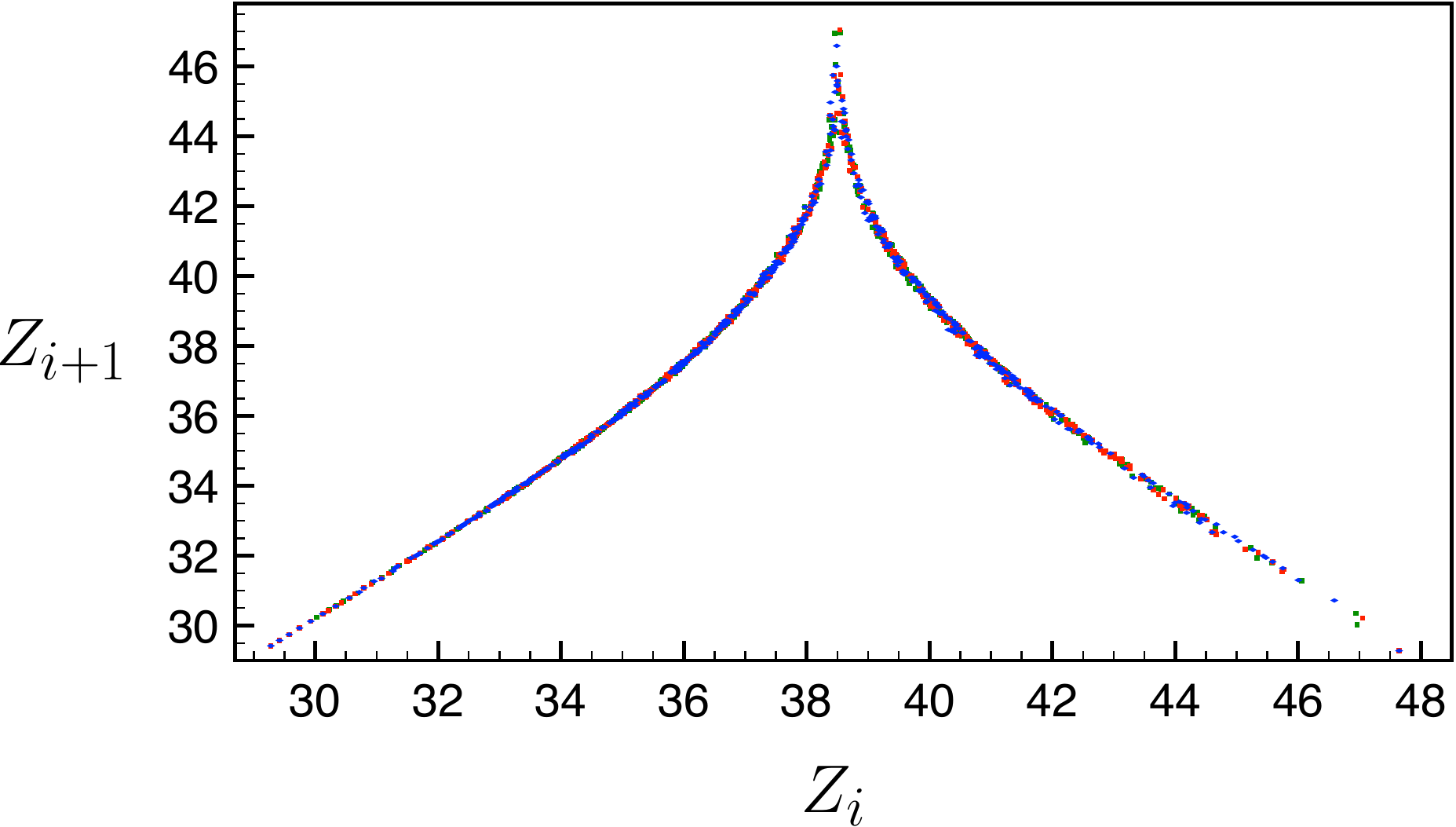}
  \end{center}  
  \caption{Comparison of return map of the Lorenz system (long term behavior) with quadruple precision (blue dots) and double (red dots) and single precision (green dots). The points are nearly superimposed revealing that the long term prediction is almost the same independently of the precision. 
}
  \label{return}
\end{figure}

However, one may ask a different type of questions. Even if the short term behavior of a specific trajectory is hard to obtain numerically in a reliable manner, is it still possible to get accurate results on statistical properties of the system for long term? To answer this question, we calculate the first return application. This application, introduced by Lorenz,  consists in plotting the successive maxima of $z (t)$ over a long period of time. For that, it is enough to locate the maxima $Z_i$ of the curve and plot the position of a given maximum $Z_{i + 1}$ as a function of the preceding one, $Z_{i}$. These data are related to the structure of the Lorenz attractor to which trajectories converge for long time. Figure \ref{return} compares such long term predictions using the RK4 algorithm to integrate \eqref{Lorenzeq} with different precision. We observe that the statistical properties at long time are not dramatically sensitive to the precision at which the calculation is performed. Even if  individual trajectories are not accurately described, their global properties are correctly described. This is similar to what distinguish climate simulations from meteorological simulations: even if individual trajectories cannot be simulated beyond a few weeks to predict the weather, long term global properties of the system (climate characteristics) can be obtained for much longer periods (years or decades).

\begin{figure}[ht]
  \begin{center}
    \includegraphics[width=0.45\textwidth]{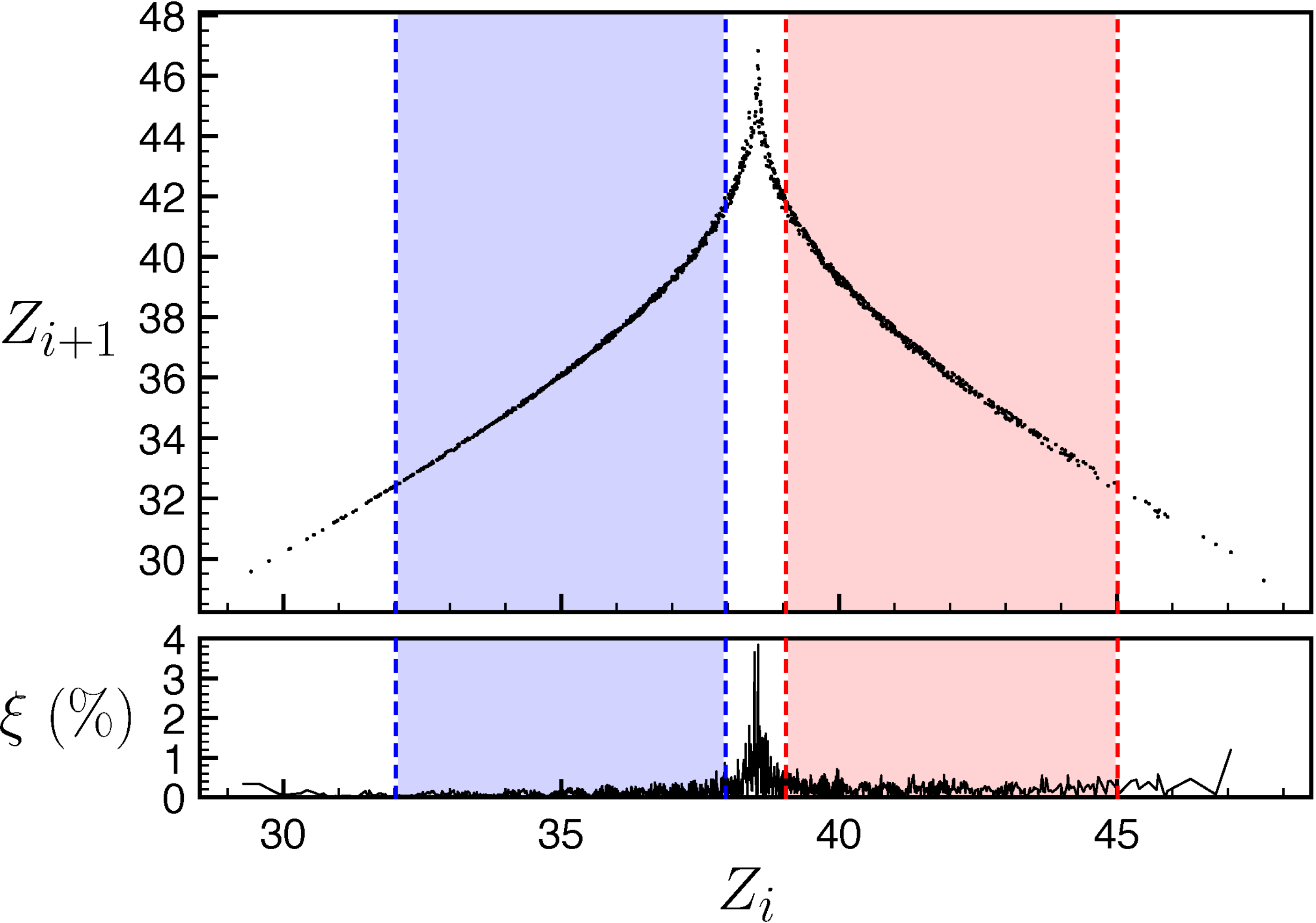}
  \end{center}  
  \caption{The return map is calculated using a RK4 integration algorithm in quadruple precision (upper panel). We fit the data in between the blue (red) dashed line with a polynomial of degree 10. We plot the relative difference $\xi$ between the fit and the return map in the lower panel.}
  \label{returnprecisiontest}
\end{figure} 

\begin{figure}[ht]
  \begin{center}
    \includegraphics[width=0.45\textwidth]{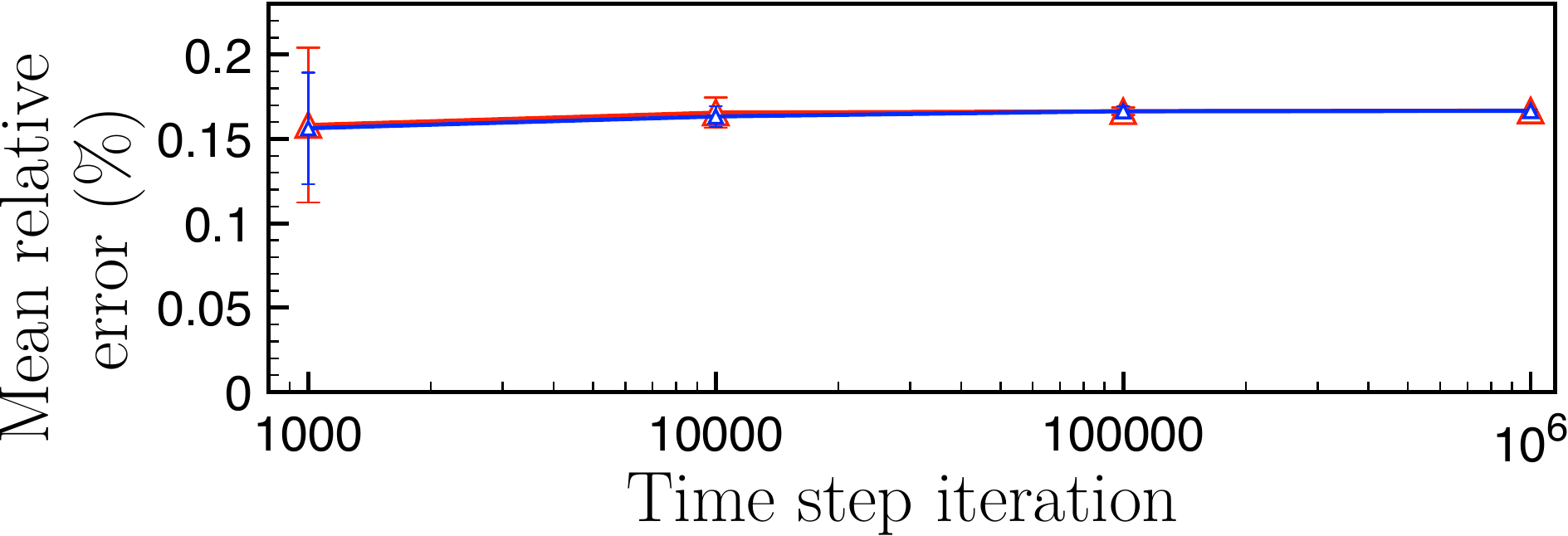}
  \end{center}  
  \caption{Mean relative error in percentage in the distance of the return map points (calculated from the RK4 algorithm) from the polynomial fit with time step $dt=0.02$ (triangle) and for single precision (large red symbol) and double precision (small blue symbol).}
  \label{returnprecisionresults}
\end{figure} 

To evaluate quantitatively the accuracy of long term simulations, we made a polynomial fit of the return map obtained with quadruple precision on each side of the peak of the return map in the window of parameters delimited by the blue dashed lines on the left side and by the red dashed lines on the right side (see  Fig.~\ref{returnprecisiontest}). We have then computed the relative error $\xi$ between the fit and the data. The mean percentage error remains below 0.2 \% in the zones delimited by the dashed line.

We then compute for single and double precision the distance towards the fit as a function of the number of iteration points considered (see Fig.~\ref{returnprecisionresults}). The results show that he mean relative error converges to less than $0.2\%$ for sufficiently large databases, in both single and double precision. The large spread of the relative error for a small number of iteration steps is due to the fact that the system has not yet reached the asymptotic behavior of the return map.  The data shown in Fig.~\ref{returnprecisionresults} indicate that the long term prediction characterized here by the return map is almost insensitive to the precision with which the trajectory is computed.
{\new We note that for this specific quantity the error cannot go to zero and has a minimum value due to the fact that the fit we use is only an approximation of the correct return map. Indeed, it is known \cite{Lorenz,Ottchaos} that the return map is actually not a curve but a fractal, of dimension slightly above one, thus close to a curve but not quite exactly. In the following, we will thus investigate in parallel as another benchmark the Lyapunov exponents of the system which have also been studied in this context \cite{Ottchaos, Ottattractor} and do not suffer from the same limitation.}

{\new In order to have data on another system, we also looked at the Lyapunov exponents of the R\"ossler system \cite{Rossler}:}

\begin{eqnarray}
\dot{x}&= -y-x \nonumber \\
\dot{y}&= x + a y \nonumber \\
\dot{z}& = b+z(x-c).
\label{Rosslereq}
\end{eqnarray}

{\new We choose the standard set of parameters: $a = 0.2, b = 0.2$ and $c = 5.7$.
This model is also chaotic and dissipative with trajectories converging to a strange attractor called the R\"ossler attractor.}

\section{Results: accuracy of predictions for the chaotic models studied}

In order to evaluate the accuracy of the machine learning approaches to predict the behavior of the chaotic systems studied, we use three different methods: a reservoir computing model as pioneered in this context in \cite{Ottchaos,OttPRL,Ottattractor}, called Echo State Network (ESN) and two other approaches based on Recurrent Neural Networks (RNN) used in \cite{LSTMchaos,3body,OttNN,Lorenzcomp}, called LSTM and TCN. The characteristics of the networks we used are detailed in the Appendix.

In this Section, we compare the predictions and performances of each network for the Lorenz system \eqref{Lorenzeq}, focusing especially on the effects of precision of both data and algorithm.  
Networks are trained on trajectories generated by the RK4  integration method and having thousands of points separated by a time step $dt = 0.02$. Predictions are performed starting immediately after the last point used in the training trajectories. Subsequent predictions are systematically done from
the point previously returned by the network.

\subsection{Resources needed for the simulation by the three neural networks}

\begin{figure}[ht]
  \begin{center}
    \includegraphics[width=0.45\textwidth]{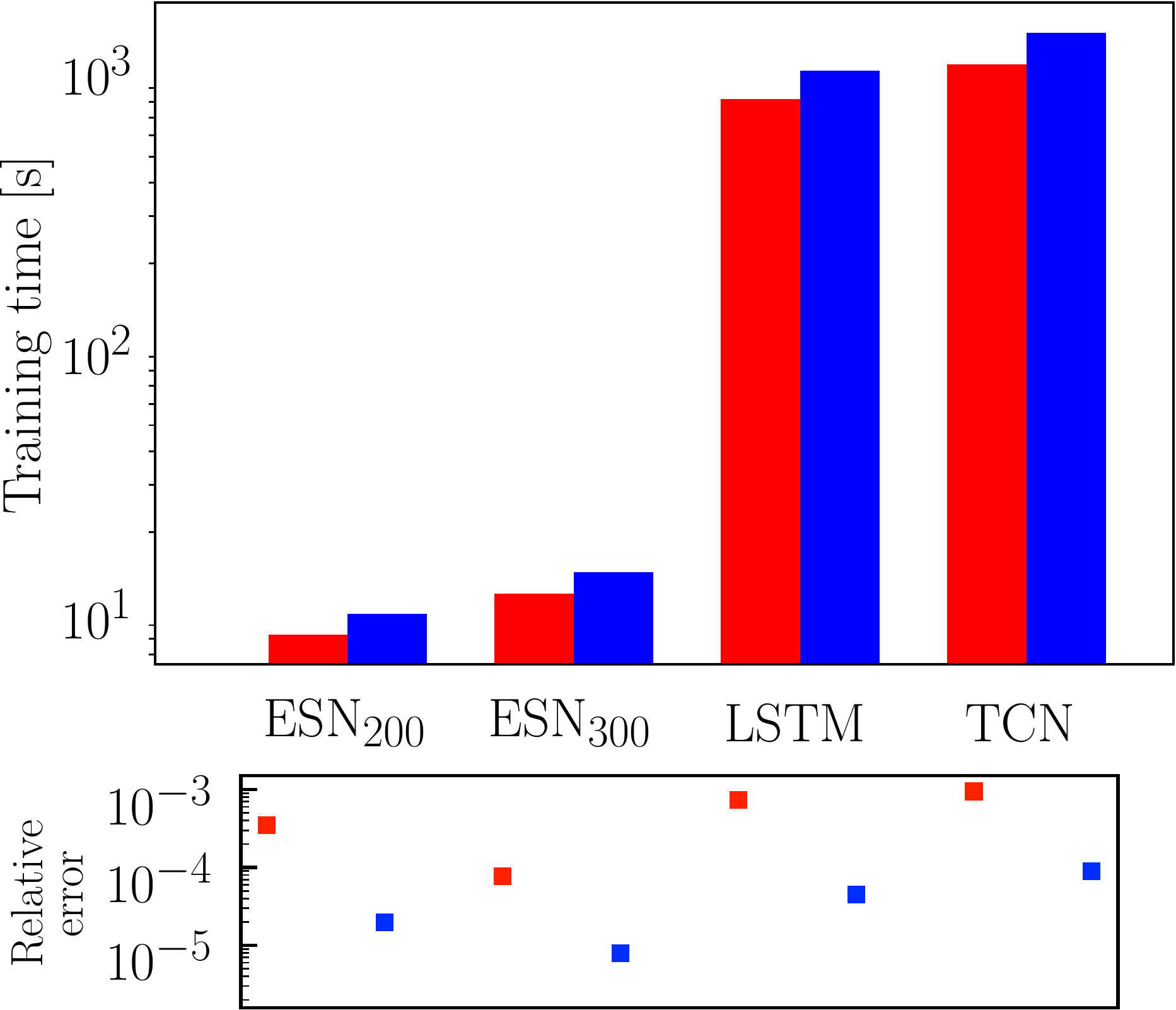}
  \end{center}  
  \caption{Upper panel: Comparison of the training time for different neural networks: ESN$_{200}$ (the reservoir contains 200 neurons),ESN$_{300}$, a LSTM network (with a single hidden layer having 64 neurons) and a TCN network. The red (blue) color is used for a computation of the networks parameters in single (double) precision. Lower panel: Figure of merit of each neural network representing the mean relative error in the estimate of the training trajectories.}
  \label{trainingtime}
\end{figure} 

\begin{figure}[ht]
  \begin{center}
    \includegraphics[width=0.45\textwidth]{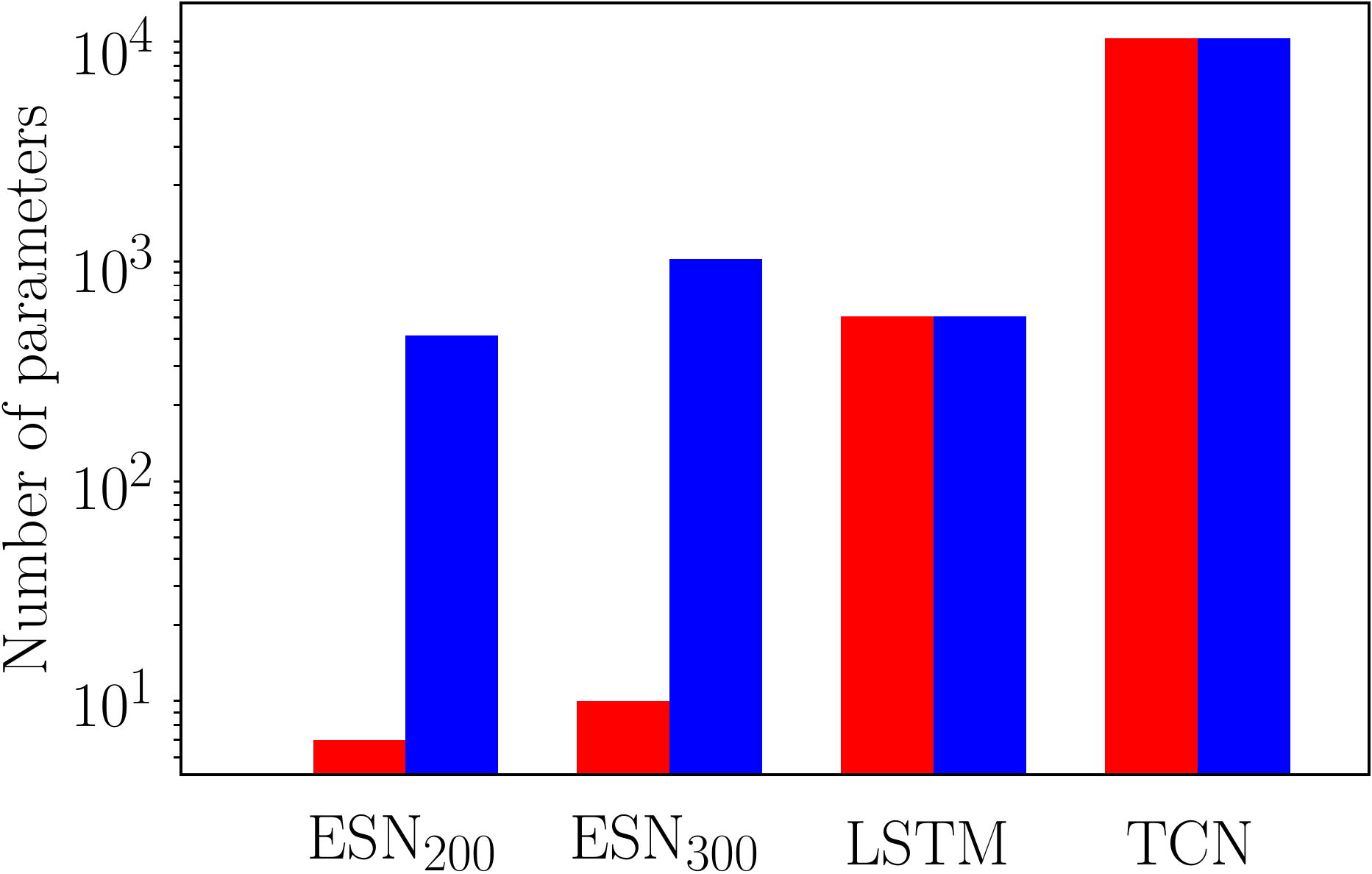}
  \end{center}  
  \caption{Comparison of number of parameters for the different neural networks considered in Fig.~\ref{trainingtime}. Red is the training size, blue the total size.}
  \label{trainingsize}
\end{figure} 

Figure~\ref{trainingtime} gives an overview of the different resources consumed during the training phase by the three networks for achieving a similar converged simulation on the same computer once the network has been set up. It is worth noticing that the performance are for a standard processor. We have not used GPU cards. For the training time, we use the same set of training trajectories (100 trajectories, each trajectory contains 50 000 points separated by a time interval $dt=0.02$).
We compare an ESN with a reservoir size having 200 neurons (ESN$_{200}$), 300 neurons (ESN$_{300}$), a LSTM network (with a single hidden layer having 64 neurons) and a TCN network (similar structure as the LSTM network). Note that the LSTM and TCN are trained 10 times on the training data set while the ESN scans the training data set only once. In addition, the number of parameters that are updated are significantly different depending on the reservoir type as illustrated in Figure.~\ref{trainingsize}. The LSTM and TCN networks adapt themselves by modifying all the network parameters. This is to be contrasted with the ESN that updates only the connections towards the output as discussed in the appendix, making the training size much smaller than the total size.

The figure of merit of each neural network is represented in the lower panel of Figure~\ref{trainingtime} where we represent the mean relative distance between the trajectories provided by the network compared to the training one. This quantity is here averaged over all the training trajectories. When this relative error is equal to 0.01, it means that the average relative error is on the order of 1 \%. As expected, for each neural network the computation of the parameters in double precision yields better results. We also see that the ESN network seems more accurate at reproducing the training trajectory. We conclude that the ESN turns out to be significantly more efficient that the LSTM and TCN networks with respect to the training time and moreover seems to better reproduce the training trajectory. 

\subsection{Short term predictions}

We now turn to the accuracy of the predictions of the different networks as compared to a quadruple precision simulation by integration of the equations of the Lorenz system \eqref{Lorenzeq}.


\begin{figure}[ht]
  \begin{center}
    \includegraphics[width=0.45\textwidth]{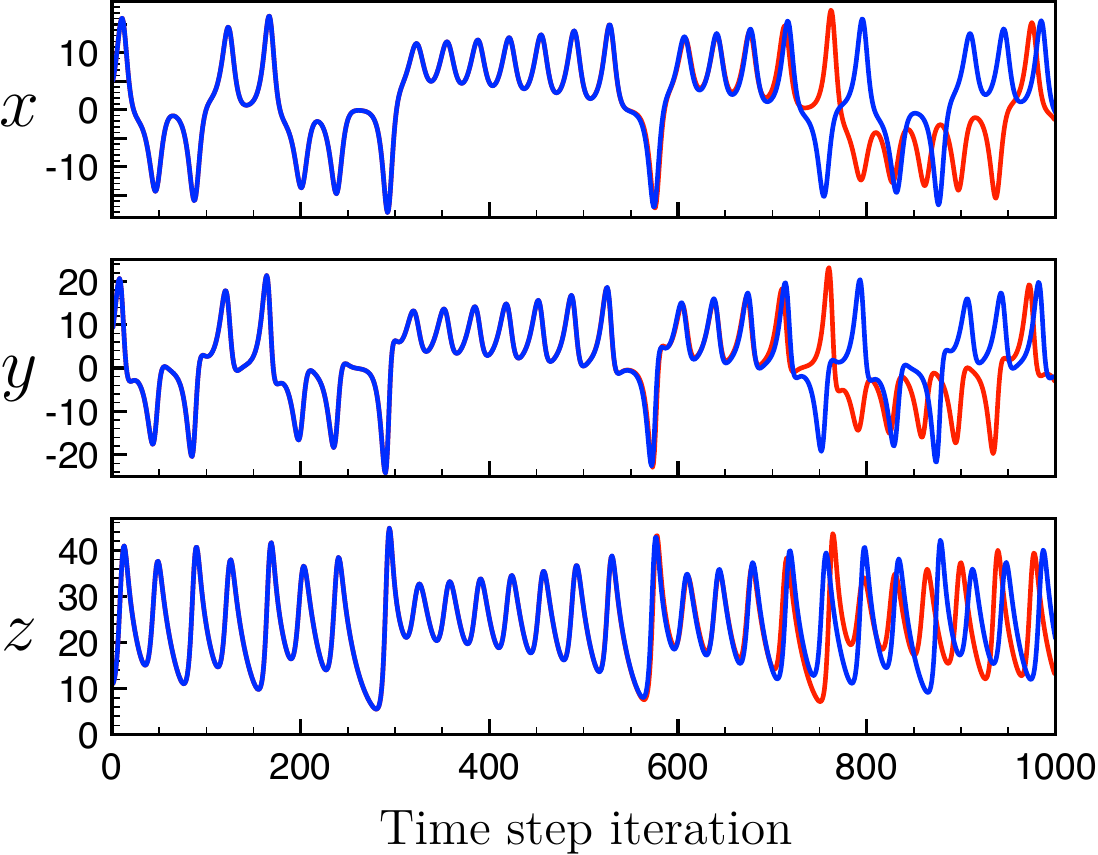}
  \end{center}  
  \caption{Comparison between the quadruple precision RK4 simulation (red line) and the prediction of the ESN in double precision with a reservoir of size $N = 300$ (blue line) for the Lorenz system \eqref{Lorenzeq}. The initial conditions are $x(0) =0.00$, $y(0) = 0.45$ and $z(0) = 1.41$, and the time step is $dt=0.02$. The ESN has been trained over 50000 time step iterations before the prediction for the subsequent iterations represented in this figure.  
  }
  \label{ESNshortex}
\end{figure}

We first look at short term predictions, i.e. accurate description of a single specific trajectory. That is the type of predictions where chaotic systems are the most difficult to handle. It is similar to meteorological predictions in weather models, since one wants a precise state of the system starting from a specific initial state.  We recall that the data
are generated via the RK4 method, with a time step of 0.02 and a sampling of
thousands of points. Our reference trajectory is calculated in
quadruple precision for the same time step and sampling.

A parameter set specific to each network architecture has been established
allowing each network to converge. They can be used to predict future points beyond the training set. As said before, the protocol is the same for the three types of networks. The output associated to input vector at time $t=T$ defines the next point for the trajectory at time $T+dt$. This procedure is iterated to get the prediction over large amount of time. We provide an example in Fig.~\ref{ESNshortex} for an ESN neural network which turns out to be able to provide an accurate prediction of the trajectory over the short term for relatively long time. 

\begin{figure}[ht]
  \begin{center}
    \includegraphics[width=0.45\textwidth]{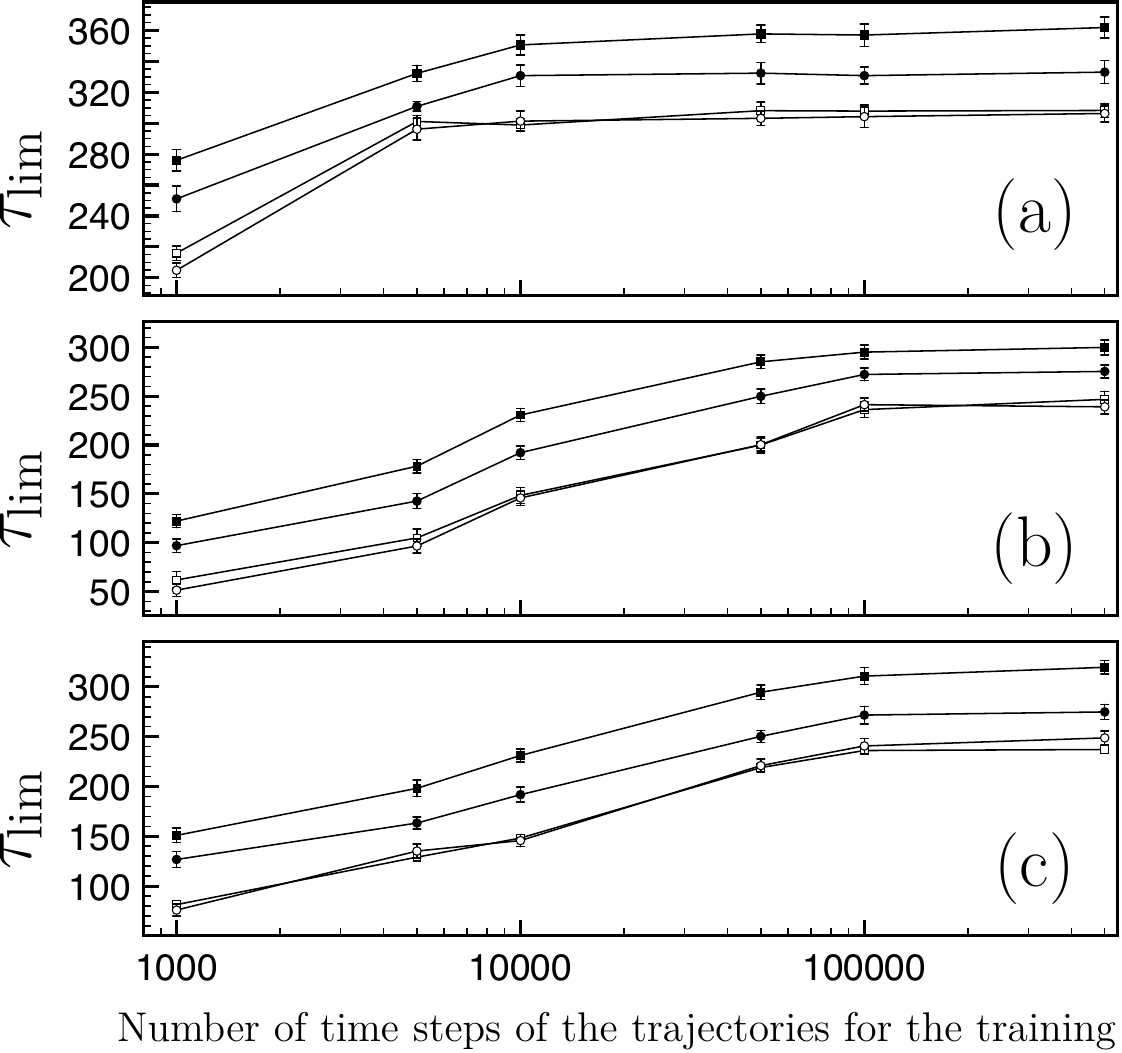}
  \end{center}  
  \caption{Impact of the precision of training data and of the neural network on the short term quantified by the time $\tau_{\rm lim}$ above which the prediction departs by more than 5\% from the quadruple precision trajectory for the Lorenz system \eqref{Lorenzeq}: (a) ESN, (b) LSTM and (c) TCN.  Data and network double precision (filled square), data single precision and network double precision (filled disk), data double precision and network single precision (empty square), data and network single precision (circle). Time step is $dt=0.02$. {\new Each point is an average over 100 simulations with randomly selected initial positions, error bars correspond to the standard deviation.}}
  \label{fig111213}
\end{figure} 

To be more quantitative, we evaluate for each simulation a limit time, $\tau_{\rm lim}$, defined as the time when the simulation departs from the correct trajectory by at least $5\%$. This quantity is plotted in Fig.~\ref{fig111213} for the three networks considered, as a function of the size of training data (number of points of the exact trajectory which are used to train the network). In all cases, one sees an increase of the limit time with increasing dataset, until it reaches a plateau where increasing the dataset does not help any more. This defines a sort of ultimate limit time for this kind of simulation.
All three neworks are effective at predicting the dynamics, giving accurate results for hundreds of time steps. The LSTM and TCN networks give very similar results, and are significantly and systematically less effective than the ESN network used in the seminal paper \cite{OttPRL}, with prediction times 20\% smaller.
We recall (see preceding subsection) that the LSTM and TCN networks are not only significantly less effective at predicting the dynamics than the ESN, they are also more costly in resources. 
The main difficulty for an ESN network is in the search for a viable parameter.  

We note that although these neural network methods are effective, they are less efficient than standard classical simulations like RK4 with lower precision (see Fig.\ref{butterfly}). We should note however that neural networks techniques are still new and far from optimized compared to integration methods. In addition, the neural network techniques do not need the equations and do not depend on approximations which can have been used to construct them.

Figure~\ref{fig111213} also enables to assess
the question of the impact of precision on the predictive abilities of the neural networks. We have changed independently the precision of the datasets used to train the network, and the precision of the network algorithm itself.  We see that in all cases the precision of the network will impact the accuracy of the prediction. Indeed, for these short term predictions, a double precision network always gives better results than a single precision network. Interestingly enough, with a single precision network, increasing the precision of the training data does not help. On the other hand, using a double precision network even on single precision data is more advantageous than a single precision network on any type of data. These results are valid for the three types of networks over the full range of training sets used.  It therefore seems that the precision of the network is crucial for the accuracy of the prediction, and more so than the precision of the data. It is especially important in view of the fact that the precision of the data can be less easily controlled than the precision of the network.

\begin{figure}[ht]
  \begin{center}
    \includegraphics[width=0.45\textwidth]{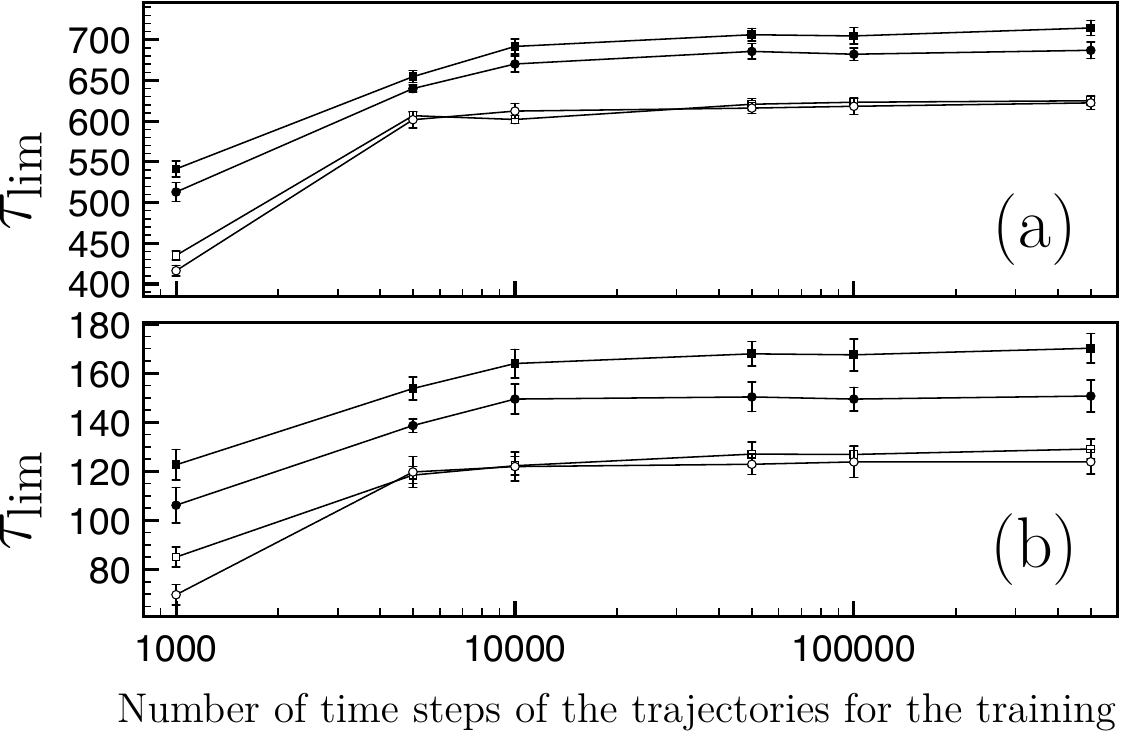}
  \end{center}  
  \caption{{\new Impact of the precision of training data and of the neural network on the short term quantified by the time $\tau_{\rm lim}$ above which the prediction departs by more than 5\% from the quadruple precision trajectory for the Lorenz system \eqref{Lorenzeq}, for different sampling times.  Data and network double precision (filled square), data single precision and network double precision (filled disk), data double precision and network single precision (empty square), data and network single precision (circle). Time step is (a) $dt=0.01$ (b) $dt=0.04$. Each point is an average over 100 simulations with randomly selected initial positions, error bars correspond to the standard deviation.}}
  \label{figdt}
\end{figure} 

{\new To verify that our results are not affected by a change of the sampling time, we display in Fig.~\ref{figdt} the same quantity as in Fig.~\ref{fig111213} for the ESN network for two additional sampling times, showing that the same result holds, the double precision network giving better results than the single precision network even when trained with lower precision data.}


\subsection{Long term predictions}


\begin{figure}[ht]
  \begin{center}
    \includegraphics[width=0.45\textwidth]{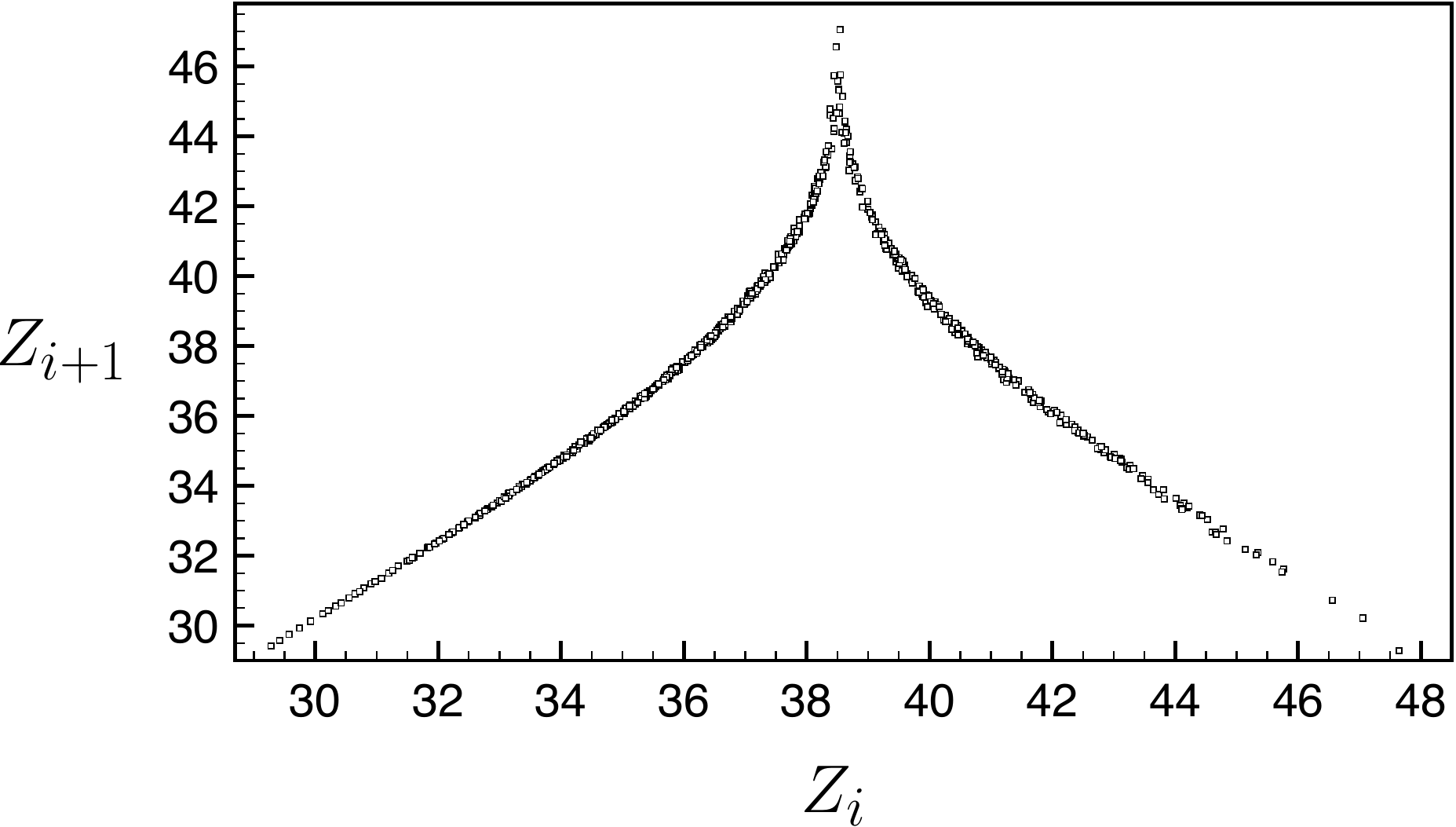}
  \end{center}  
  \caption{Return map of the Lorenz system obtained by an ESN network simulation.}
  \label{ESNreturnmap}
  \end{figure} 

\begin{figure}[ht]
  \begin{center}
    \includegraphics[width=0.45\textwidth]{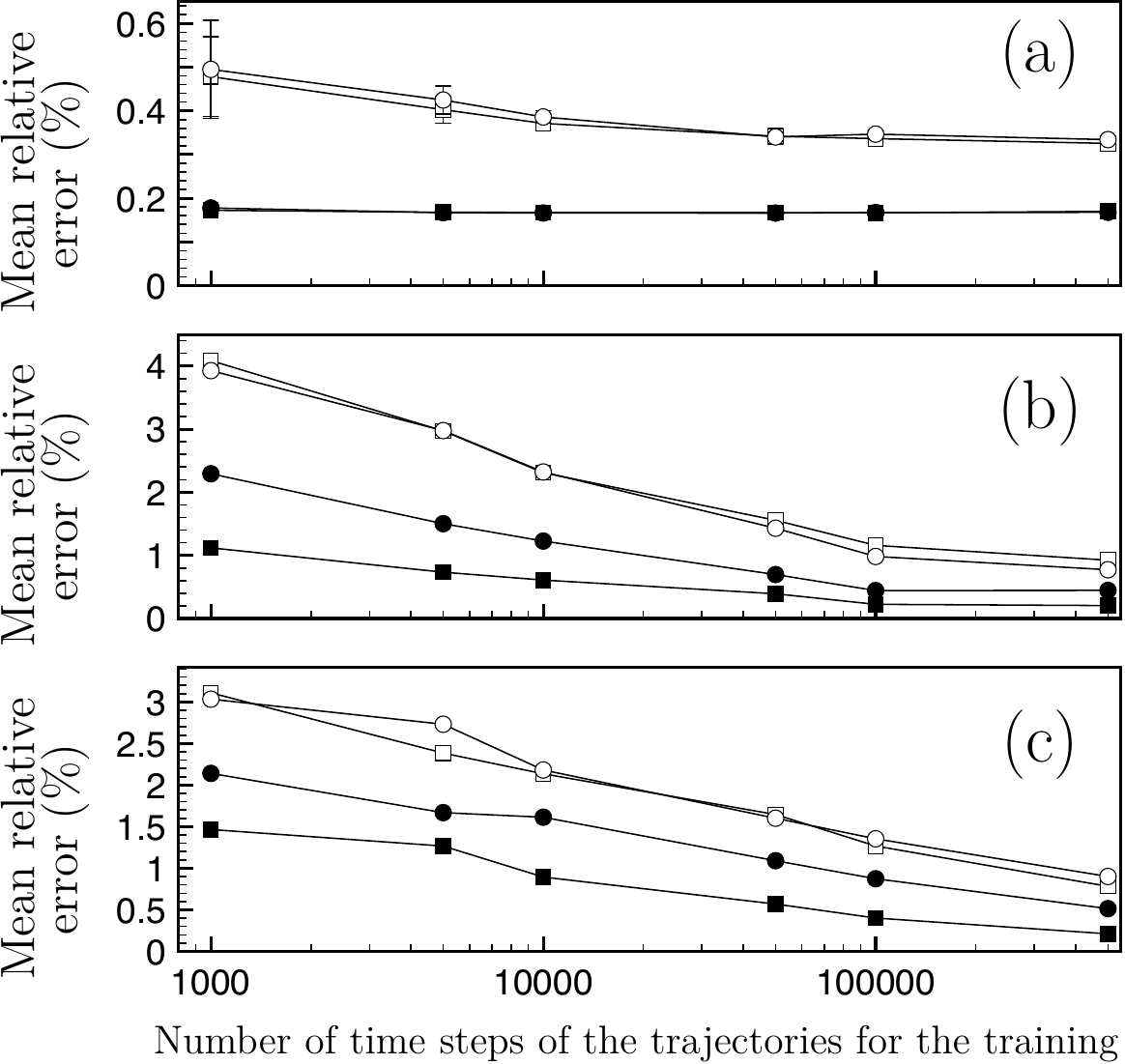}
  \end{center}  
  \caption{{\new Impact of the precision of training data and of the neural network on the long term quantified by the mean relative error defined in Section II for the return map of the Lorenz system \eqref{Lorenzeq}: (a) ESN, (b) LSTM and (c) TCN.  Data and network double precision (filled square), data single precision and network double precision (filled disk), data double precision and network single precision (empty square), data and network single precision (circle). Time step is $dt=0.02$. Each point is an average over 100 simulations with randomly selected initial positions, error bars correspond to the standard deviation.}}
  \label{fig161718}
\end{figure} 

We now turn to long term predictions.
{\new Figure \ref{ESNreturnmap} displays an example of return map constructed from ESN predictions, showing that, despite the fact that specific individual trajectories are not accurately simulated for long times, nevertheless the points predicted by the network describe correctly the long time dynamics of a typical trajectory, giving the general shape of the return map.}

To be more quantitative, Fig.~\ref{fig161718} uses the measures defined in Section II (see Fig.~\ref{returnprecisiontest}) to assess the efficiency of the neural network methods for long term dynamics. Despite the fact that the LSTM and TCN networks are more cumbersome to implement and take more running time, the results are clearly better for the ESN network, which can achieve an accuracy similar to the one of the RK4 simulations (see Fig.~\ref{fig161718} a). For the LSTM and TCN networks, the results presented in Figs.~\ref{fig161718} b) and c) show that these networks are able to reproduce the long term dynamics, but the accuracy is less good than for ESN networks or RK4, even for large sizes of the training dataset.

\begin{table}
\begin{tabular}{ |c|c|c|c|c| } 
 \hline
 	   	     & RK4      & ESN         & LSTM  		& TCN \\ 
 \hline
 $\Lambda_1$ & 0.906    & 0.884  $\pm 10^{-2}$    & 0.871 	$\pm 2.10^{-2}$ 	& 0.879$\pm 2.10^{-2}$ \\ 
 $\Lambda_2$ & 0        & $3 \; 10^{-5}$ $\pm 10^{-4}$& $-1.10^{-5}$ $\pm 2.10^{-4}$ & $-1.10^{-5}$ $\pm 2.10^{-4}$\\ 
 $\Lambda_3$ & -14.567  & -9.203  $\pm 0.2$    & -8.498 	$\pm 0.3$ 	& -8.614 $\pm 0.3$  \\ 
 \hline
\end{tabular}
\caption{{\new Lyapunov spectrum of the Lorenz system \eqref{Lorenzeq} estimated by long time evolution through the different types of network. Each network is trained with 10000 points on the Lorenz system and predicts 50000 points. Here, the networks are in single precision and the data are in single precision. Each point is an average over 100 simulations with randomly selected initial positions, error bars correspond to the standard deviation.}}
\label{table1}
\end{table}

\begin{table}
\begin{tabular}{ |c|c|c|c|c| } 
 \hline
 	   	     & RK4      & ESN         & LSTM  		& TCN \\ 
 \hline
 $\Lambda_1$ & 0.906    & 0.891    $\pm 10^{-2}$    & 0.872 	$\pm 2.10^{-2}$	& 0.885 $\pm 2.10^{-2}$\\ 
 $\Lambda_2$ & 0        & $-1.10^{-5}$ $\pm 10^{-4}$ & $1.10^{-5}$ $\pm 2.10^{-4}$& $-2.10^{-5}$ $\pm 2.10^{-4}$\\ 
 $\Lambda_3$ & -14.567  & -9.471    $\pm 0.3$     & -8.566 	$\pm 0.3$ 	& -8.681 $\pm 0.3$ \\ 
 \hline
\end{tabular}
\caption{{\new Lyapunov spectrum of the Lorenz system \eqref{Lorenzeq} estimated by long time evolution through the different types of network. Each network is trained with 10000 points on the Lorenz system and predicts 50000 points. Here, the networks are in single precision and the data are in double precision. Each point is an average over 100 simulations with randomly selected initial positions, error bars correspond to the standard deviation.}}
\label{table2}
\end{table}

\begin{table}
\begin{tabular}{ |c|c|c|c|c| } 
 \hline
 	   	     & RK4      & ESN          & LSTM  		  & TCN \\ 
 \hline
 $\Lambda_1$ & 0.906    & 0.896    $\pm 10^{-2}$      & 0.881 	$\pm 10^{-2}$ 	  & 0.894 $\pm 10^{-2}$\\ 
 $\Lambda_2$ & 0        & $-6.10^{-6}$ $\pm 10^{-4}$ & $-4.10^{-6}$ $\pm 2.10^{-4}$& $-5.10^{-6}$ $\pm 2.10^{-4}$\\ 
 $\Lambda_3$ & -14.567  & -10.102   $\pm 0.3$    & -9.345	$\pm 0.3$ 	  & -9.460 $\pm 0.3$\\ 
 \hline
\end{tabular}
\caption{{\new Lyapunov spectrum of the Lorenz system \eqref{Lorenzeq} estimated by long time evolution through the different types of network. Each network is trained with 10000 points on the Lorenz system and predicts 50000 points. Here, the networks are in double precision and the data are in single precision. Each point is an average over 100 simulations with randomly selected initial positions, error bars correspond to the standard deviation.}}
\label{table3}
\end{table}

\begin{table}
\begin{tabular}{ |c|c|c|c|c| } 
 \hline
 	   	     & RK4      & ESN           & LSTM  		  & TCN \\ 
 \hline
 $\Lambda_1$ & 0.906    & 0.900    $\pm 3.10^{-3}$       & 0.885	 $\pm 8.10^{-3}$ 	  & 0.894 $\pm 9.10^{-3}$ \\ 
 $\Lambda_2$ & 0        & $-5.10^{-6}$ $\pm 10^{-4}$ & $-4.10^{-5}$ $\pm 2.10^{-4}$& $-4.10^{-5}$ $\pm 2.10^{-4}$\\ 
 $\Lambda_3$ & -14.567  & -10.985  $\pm 0.3$     & -9.513 $\pm 0.3$ 		  & -9.629 $\pm 0.3$\\ 
 \hline
\end{tabular}
\caption{{\new Lyapunov spectrum of the Lorenz system \eqref{Lorenzeq} estimated by long time evolution through the different types of network. Each network is trained with 10000 points on the Lorenz system and predicts 50000 points. Here, the networks are in double precision and the data are in double precision. Each point is an average over 100 simulations with randomly selected initial positions, error bars correspond to the standard deviation.}}
\label{table4}
\end{table}

{\new The data for the Lyapunov exponents of the Lorenz system \eqref{Lorenzeq} are shown in the Tables \ref{table1}-\ref{table4}. They display the same trend as the data for the return map. Again, the ESN network performs best, giving the best approximation of the Lyapunov exponents wherever the difference is significative. The double precision network is better at approximating the correct Lyapunov, even if it is trained with single precision data. On the contrary, training with double precision data a single precision network gives clearly worse results. We note that the third Lyapunov is less well approximated by the network, as was noted already in \cite{Ottchaos}, since it is associated with the deviation of the return map from the approximating curve. }

\begin{table}
\begin{tabular}{ |c|c|c|c|c|c| } 
 \hline
 	   	     & RK4      &$ESN_{sp-sp}$ &$ESN_{sp-dp}$    \\ 
 \hline
 $\Lambda_1$ & 0.067    & 0.062  $\pm 3.10^{-3}$       & 0.063	$\pm 3.10^{-3}$	  \\ 
 $\Lambda_2$ & 0        & $-3.10^{-6}$ $\pm 2.10^{-4}$ & $3.10^{-5}$ $\pm 2.10^{-4}$  \\ 
 $\Lambda_3$ & -5.41    & -4.377  $\pm 0.2$      & -4.434	$\pm 2.10^{-3}$	   \\ 
 \hline
 	   	     & RK4       &$ESN_{dp-sp}$   &$ESN_{dp-dp}$ \\ 
 \hline
 $\Lambda_1$ & 0.067   	   & 0.066  $\pm 2.10^{-3}$        & 0.066 $\pm 2.10^{-3}$\\ 
 $\Lambda_2$ & 0        & $-8.10^{-7}$ $\pm 2.10^{-4}$  & $-7.10^{-6}$ $\pm 2.10^{-4}$\\ 
 $\Lambda_3$ & -5.41      & -4.797  $\pm 0.1$       & -4.851 $\pm 9.10^{-2}$ \\ 
 \hline
\end{tabular}
\caption{{\new Lyapunov spectrum for the R\"ossler system \eqref{Rosslereq} with $a = 0.2, b = 0.2, c = 5.7$ estimated by long time evolution through the ESN network (reservoir computing), with different precisions: from left to right and top to bottom network and data single precision, network single precision and data double precision, network double precision and data single precision, network double precision and data double precision. The network is trained with 10000 points and predicts 50000 points. Each point is an average over 100 simulations with randomly selected initial positions, error bars correspond to the standard deviation.}}
\label{table5}
\end{table}

{\new We also include data on the the computation of the Lyapunov exponents for the R\"ossler system of \eqref{Rosslereq} in Table \ref{table5}, using only the ESN (reservoir computing) network. Again, even if the third Lyapunov is not well approximated as for the Lorenz system, the results indicate that the double precision network is better in all cases, independently of the precision of the training data.}

\begin{table}
\begin{tabular}{ |c|c|c|} 
 \hline
 	   	        & mean     & standard deviation \\ 
 \hline
 $error_{sp\_sp}$ & 0.009327 & 0.001065 \\ 
 $error_{sp\_dp}$ & 0.007148 & 0.000934 \\ 
 $error_{dp\_sp}$ & 0.000378 & 0.000121 \\ 
 $error_{dp\_dp}$ & 0.000359 & 0.000107 \\ 
 \hline
\end{tabular}
\caption{{\new Long time mean error and standard deviation of predicted trajectories of the Lorenz system \eqref{Lorenzeq} for the ESN network, for different precisions: from top to bottom network and data single precision, network single precision and data double precision, network double precision and data single precision, network double precision and data double precision. Starting from the 25000th predicted point, we compare the point at timestep $t+1$ from the ESN trajectories to the one obtained by integrating \eqref{Lorenzeq} through RK4 from the predicted point at timestep $t$. We repeat this procedure for 10000 points. Each error is further averaged over 100 trajectories with random initial conditions.}}
\label{table6}
\end{table}

{\new At last, we include a different test on long time which verifies how the local dynamics at long times is accurately simulated. Starting from a given time, we compare the points predicted by the network to the integration of the equation of the Lorenz system \eqref{Lorenzeq} starting from the previously predicted point, and average the error obtained over many consecutive points. The results for the ESN network are displayed on Table~\ref{table6}, and again the double precision network gives an error an order of magnitude smaller than the single precision network, almost independently of the precision of the data.}

{\new Globally, as in the case of short term predictions, the results presented in Figs.~\ref{fig161718} and Tables \ref{table1}-\ref{table6} allow us to estimate the effects of the precision on long term predictions. The ESN network fares better in predicting the correct quantities, and in all cases it is clear that the precision of the results is dominantly controlled by the precision of the network, independently of the precision of the training data: even with low precision data, the high precision network fares better than a low precision network with high precision data.}



\section{Conclusion}

The results presented in this paper confirm previous works, showing that neural networks are able to simulate chaotic systems, both for short term and long term predictions. We also show that the ESN network (reservoir computing) seems globally more efficient in this task than LSTM or TCN networks, in line with the recent work \cite{OttNN}. Our investigations allow to assess the effect of the precision of the training data and precision of the network on the accuracy of the results. Our results show than in a very consistent manner, the precision of the network matters more than the precision of the data on which it is trained. {\new It may seem surprising that using exactly the same single precision data, changing the precision of the algorithm can give results much closer to a full double precision simulation.} {\new However, this is good news for practical applications, such as meteorology or climate simulations. Indeed, eventhough the errors in the training datasets considered in this paper are not observational errors, and are due solely to the precision of the integration of the equations, our results seem to indicate that one can compensate the lack of precision of the dataset by increasing the precision of the network. In many practical instances, the precision of the datasets is given by the precision of observations, that may be hard to ameliorate, while the precision of the network is controlled at the level of the algorithm used and may be increased at a cost of more computing time.}

\begin{acknowledgments}
We thank Gael Reinaudi for discussions. We thank CalMiP for access to its supercomputer.  This study has been supported through the EUR grant NanoX  ANR-17-EURE-0009 in the framework of the ``Programme des Investissements d'Avenir''.
\end{acknowledgments}

\appendix

{

\section{The three machine learning approaches used}

In the Appendix, we give an overview of the main features of the three neural networks that have been used in the article, namely the ESN, LSTM and TCN networks.

\subsection{Reservoir computing: ESN network}

\begin{figure}[ht]
  \begin{center}
    \includegraphics[width=0.45\textwidth]{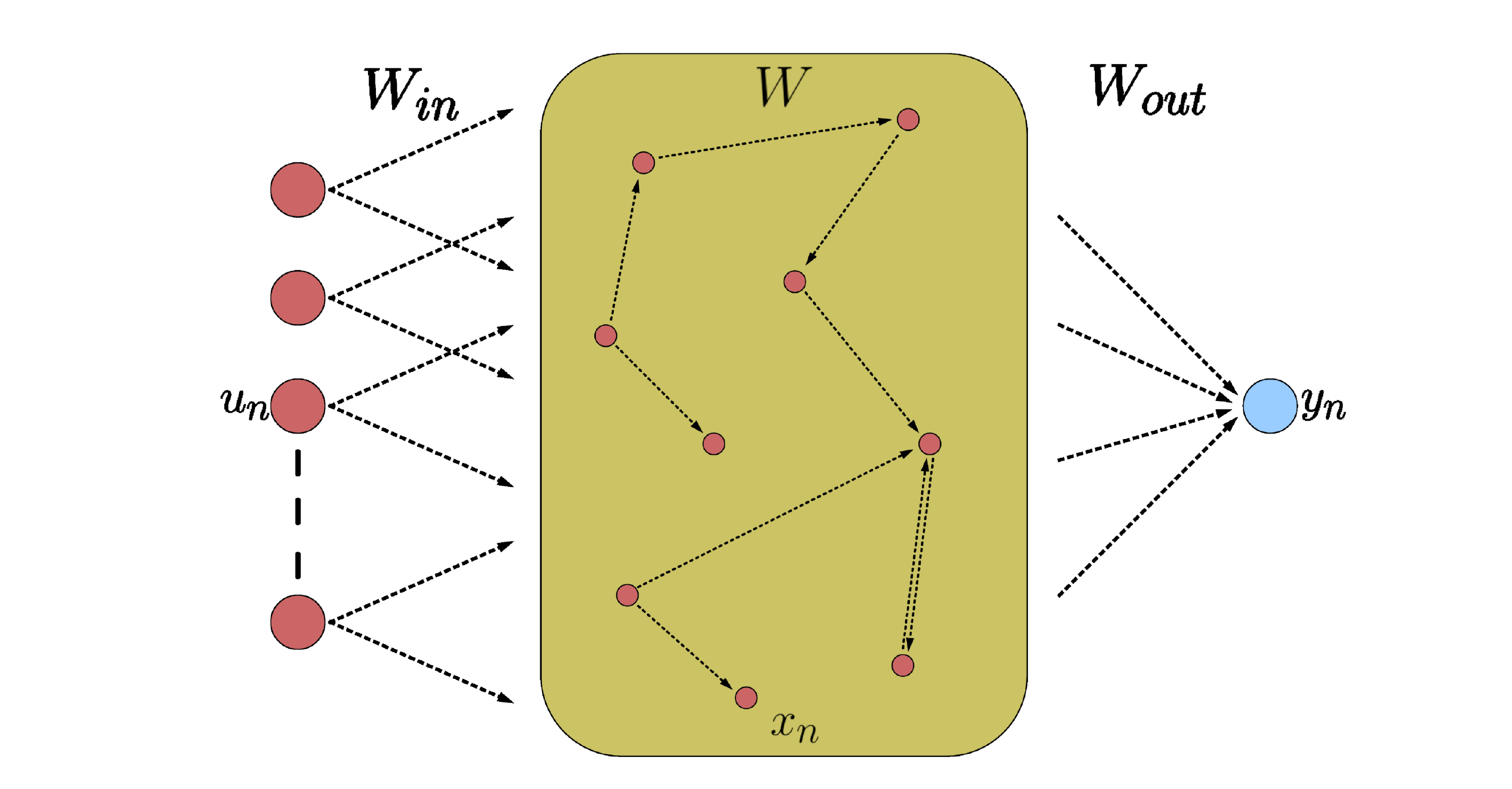}
  \end{center}  
  \caption{Schematic representation of an Echo State Network (ESN).
}
  \label{ESN}
\end{figure} 

The first network we use corresponds to reservoir computing. We focus on a specific model called Echo State Networks (ESN).  Reservoir computing methods were developed in the context of computational neuroscience  to mimick the processes of the brain. Their success in machine learning comes from the fact that they are relatively cheap in computing time and have a simple structure. Their complexities lie in their training and the adjustment of parameters to obtain the desired results.
The structure of ESN networks is schematized on Fig. \ref{ESN}. 

To train our ESN on a time-dependent signal $u_n$ with $n=1,...,T$ where $T$ is the duration of the sequence in discretized time, we must minimize a cost function between $y^{ref}_n$ and $y_n$. Here $y^{ref}_n$ is the output that we want to obtain with $u_n$, and $y_n$ is the output of the network when we give it $u_n$ as input. 
For the Lorenz problem, $u_n$, $y^{ref}_n$ and $y_n$ are 3D vector.
Generally, the cost function that one seeks to minimize is the error between the output of the network and the reference signal. This function is often in the form of a mean square error or, in our case, of the mean standard deviation.

The output of the network is calculated as follows:
\begin{equation}
y_n = W_{out} [1; u_n; x_n], 
\end{equation}
where $W_{out}$ is the output weight matrix that we are trying to train, [.;.] represents the concatenation, $u_n$ is our vectorial input signal and $x_n$ the vector corresponding to the reservoir neuron activations. It has the dimension $N$ of the reservoir and is calculated as follows:
\begin{equation}
x_n = (1-\alpha)x_{n-1} + \alpha \tilde{x}_n 
\end{equation}
with $\tilde{x}_n$ corresponding to the new value of $x_n$ :
\begin{equation}
\tilde{x}_n =  \mbox{tanh}(W_{in}[1;u_n] + Wx_{n-1} + \varepsilon_0 + \mu_0)
\end{equation}
$\alpha$ is the leaking rate, $\varepsilon_0=-1.154$ is an offset optimized on our set of data, $\mu_0$ is a random Gaussian variable of standard deviation equal to $2.25 \times 10^{-5}$, $W$ is the system reservoir and $W_{in}$ is the input weight matrix of the reservoir.
The dimension of $W_{in}$ is $N\times (3+3)$ the $+3$ term accounts for the bias added to the input (see Fig.~\ref{ESN}). The initialization consists in setting $x$ and $y$ to zero.

There are several important parameters that must be adjusted depending on the problem we are studying if we want our ESN to be able to predict our system.
The first parameter we can play on is the size of the reservoir itself. The more complex the problem that we want to deal with, the more the size of the reservoir will have an impact on the capacities of the network. A large reservoir will generally give better results than a small reservoir.
Once the size of our reservoir has been chosen, we can play on the central parameter of an ESN: the spectral radius of the reservoir. Often denoted by $\rho (W)$, this is the maximal absolute value of eigenvalues of the matrix $W$. The spectral radius determines how quickly the influence of an input data dissipates in the reservoir over time.
If the problem being treated has no long-term dependency, there is no need to keep data sent far in advance. We can therefore ensure that the spectral radius is unitary. In some cases, if our problem has long-term dependencies, it is possible to have $\rho(W)> 1$ to keep the data sent in the network longer.
The last parameter we can play on is the leaking rate $\alpha$. It characterizes the speed at which we come to update our reservoir with the new data that we provide over time.

 
 The matrices $W$ and $W_{in}$ are initialized at the start but are not modified during training.
Only the output matrix $W_{out}$ is driven:
\begin{equation}
W_{out} = Y^{ref}X^T(XX^T + \beta I)^{-1}
\end{equation}
where, for our Lorenz problem, $X=(x_1, ..., x_T)$ (dimension $N\times T$), $Y^{ref}=(y^{ref}_1, ..., y^{ref}_T)$ (dimension $3\times T$) and $I$ is the $N\times N$ identity matrix. As a result, the dimension of $W_{out}$ is $3\times N + 4$.

The fact that connectivities from input to hidden layer and from hidden to hidden layer are fixed and randomly assigned from the beginning reduces considerably the number of parameters to be trained. As a result, the training speed of the network is small compared to other networks specialized in learning specific temporal patterns (see below). 
By increasing the size of the training data, the network becomes more sensitive to the small fluctuations that accumulate during $W_{out}$ calculation. The parameter $\beta$ makes it possible to limit this dependence by penalizing the too large values of $W_{out}$. This is all the more true for a single precision network which is more sensitive to these fluctuations and whose $\beta$ must vary by several orders of magnitude depending on the size of the training data. In double precision (float64), $\beta$ varies from $10^{-8}$ for $5000$ training points to $10^{-7}$ for $5.10^5$ training points against $10^{-4}$ to $10^{-1}$ in single precision (float32).
As the reservoir is not changed during training, one must choose the initialization hyperparameters to ensure that one has a consistent output with the expected results. This requires adjusting the values of the leaking rate, spectral radius and input scaling as a priority. The optimization of these parameters has been done via a grid search where we decrease the search step as we find good parameters.

The initialization parameters are for $W_{in}$ a density equal to $d=0.464$ with values randomly distributed from a Gaussian function of  standard deviation equal to $\sigma=3.352$. For the reservoir matrix $W$, we have chosen $d_W=0.483$, $\sigma_W=2.901$, and a spectral radius $\rho_W=0.625$. 

\subsection{LSTM and TCN networks}

\begin{figure}[ht]
  \begin{center}
    \includegraphics[width=0.45\textwidth]{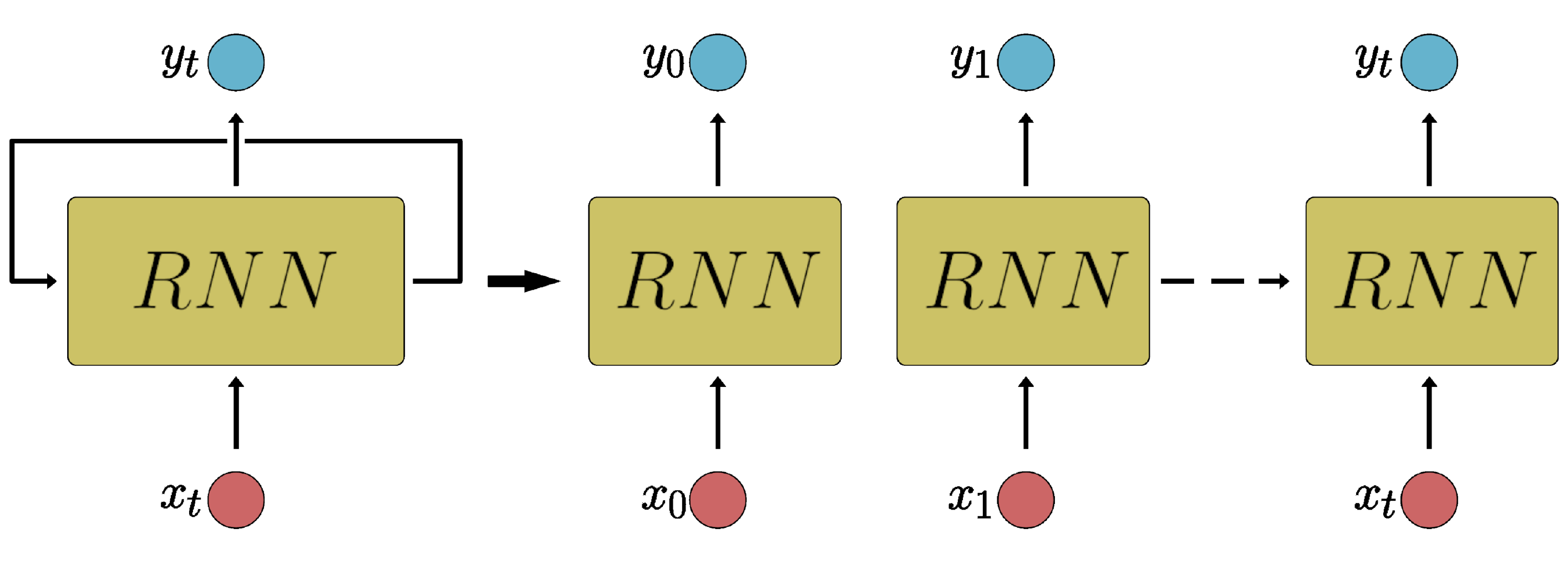}
  \end{center}  
  \caption{General structure of Recurrent Neural Networks (RNN).
}
  \label{RNN}
\end{figure}

The two other networks we use are based on
Recurrent Neural Network architectures (RNN). 
RNNs can be represented as a single module chain (see Fig. \ref{RNN}). The length of this chain depends on the length of the sequence that is sent to the input. The output of the previous module serves as input for the next module in addition to the data on which we train our network. This allows the network to keep in memory what has been sent previously.

The major problem in this type of network is the exponential decrease of the gradient during the training of the network. This is due to the nature of back-propagation of the error in the network. The gradient is the value calculated to adjust the weights in the network, allowing the network to learn. The larger the gradient, the greater the adjustments in the weights, and vice versa. When applying back-propagation to the network, each layer calculates its gradient from the effect of the gradient in the previous layer. If the gradient in the previous layer is small, then the gradient in the current layer will be even smaller. The first layers in the network therefore have almost no de facto adjustment in their weight matrices for a very small gradient.

\begin{figure}[ht]
  \begin{center}
    \includegraphics[width=0.45\textwidth]{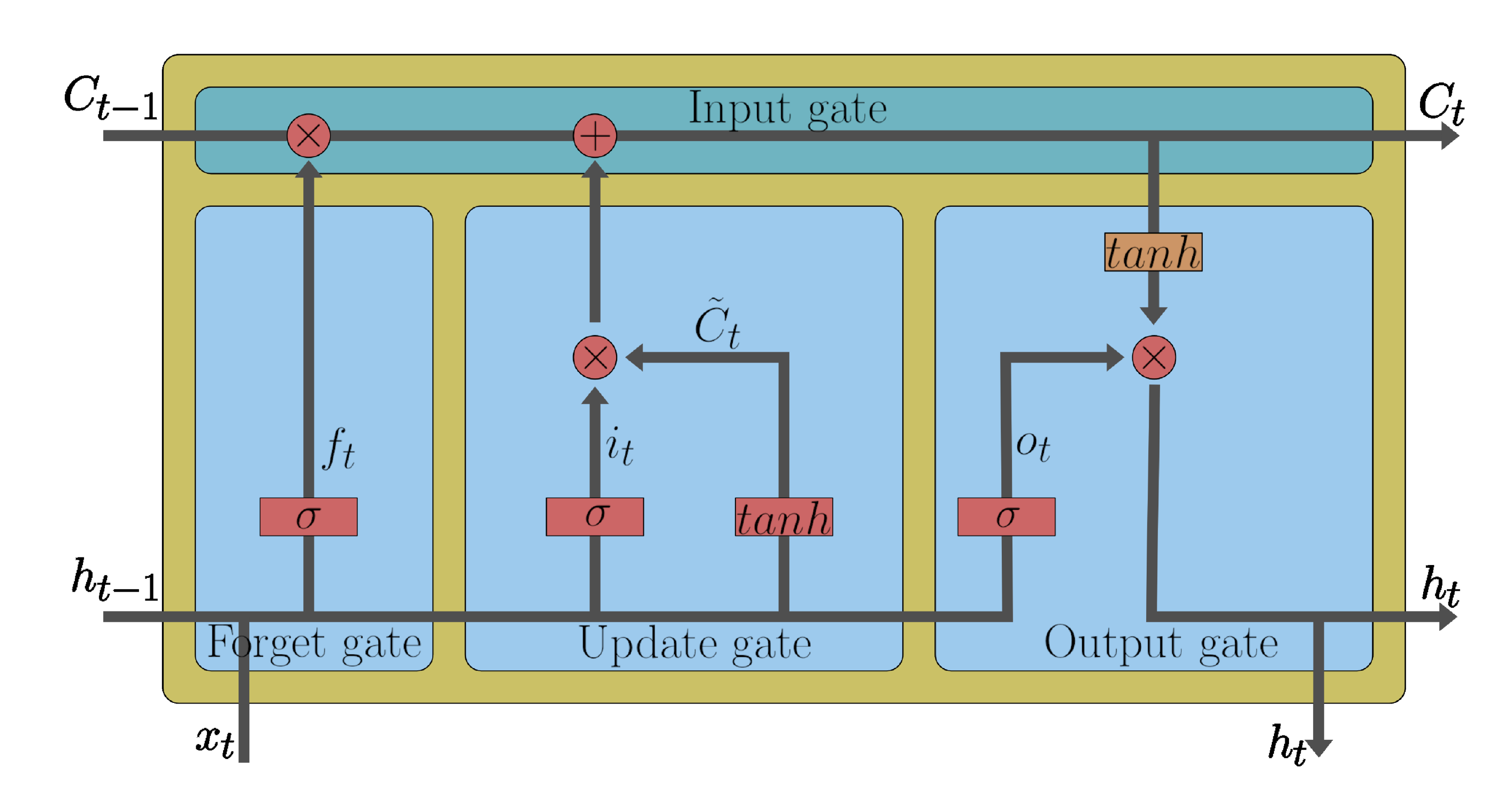}    
  \end{center}  
  \caption{Schematic representation of a Long Short Term Memory (LSTM) network: structure of one elementary cell.
}
  \label{LSTM}
\end{figure} 

To solve this problem of attenuation of the corrections, the LSTM networks (Long Short Term Memory networks) have been explicitly developed for this purpose. They can also be represented as a module chain, but unlike conventional RNNs, they have a more complex internal structure, composed of four layers which interact with each other (see Fig. \ref{LSTM}).

The first layer is called ``input gate'', and is represented by a horizontal line that runs through the entire cell. It allows data to easily browse the entire network. This structure represents the state of the cell over time. On this line there are other structures which will be used to modify the data which go through the cell.

The next step in our network is the forget gate structure. It consists of a neural network with an activation function of the sigmoid type and makes it possible to decide which part will be kept in the cell:
\begin{equation}
f_t = \sigma(W_f[h_{t-1},u_t] + b_f),
\end{equation}
where $W_f$ and $b_f$ are the weights and bias of the network for the update gate layer, $u_t$ is the input data at time $t$ and $h_{t-1}$ is the hidden state output by the previous cell.

The second step is to decide what to store. This structure consists of two parts. The first part is a neural network with an activation function of the sigmoid type, and will allow us to choose which value will be updated in our structure:
\begin{equation}
i_t = \sigma(W_i[h_{t-1},u_t] + b_i),
\end{equation}
where $W_i$ and $b_i$ are the weights and bias of the sigmoid network for the update gate layer. $u_t$ is the input data at time step $t$ and $h_{t-1}$ is the hidden state output by the previous cell.
The second part is another neural network with this time an activation function of the hyperbolic tangent type and that allows to create the new state candidate of our vector $C_t$ as follows:
\begin{equation}
\tilde{C}_t = \mbox{tanh}(W_c[h_{t-1},u_t] + b_c),
\end{equation}
where $W_c$ and $b_c$ are the weights and bias of the hyperbolic tangent network for the update gate layer, $u_t$ is the input data at time $t$ and $h_{t-1}$ is the hidden state output by the previous cell.
The new cell state $C_t$ is then computed by adding the different outputs from the input gate, the forget gate and the update gate as follows:
\begin{equation}
C_t = f_t*C_{t-1} + i_t*\tilde{C}_t,
\end{equation}
where $f_t$ is the output of the forget gate layer, $i_t$ is the input layer choosing which values are going to be updated, $\tilde{C}_t$ is the new cell state candidate, $C_{t-1}$ is the cell state of the previous cell and $*$ denote elementwise operation.
The structure described above is then repeated from cell to cell.

A final structure makes it possible to determine what will be the output of the network. The output will be based on the state of the cell to which we have applied a network with a sigmoid activation function to choose which part will be returned. Then we apply a hyberbolic tangent function to the cell state and multiply it with the previous value to get the new cell state output:
\begin{align}
o_t &= \sigma(W_o[h_{t-1},u_t] + b_o), \\
h_t &= o_t * \mbox{tanh}(C_t).
\end{align}

$h_t$ is then sent into a linear layer for prediction of $y_t$.

\begin{figure}
  \begin{center}
    \includegraphics[width=0.45\textwidth]{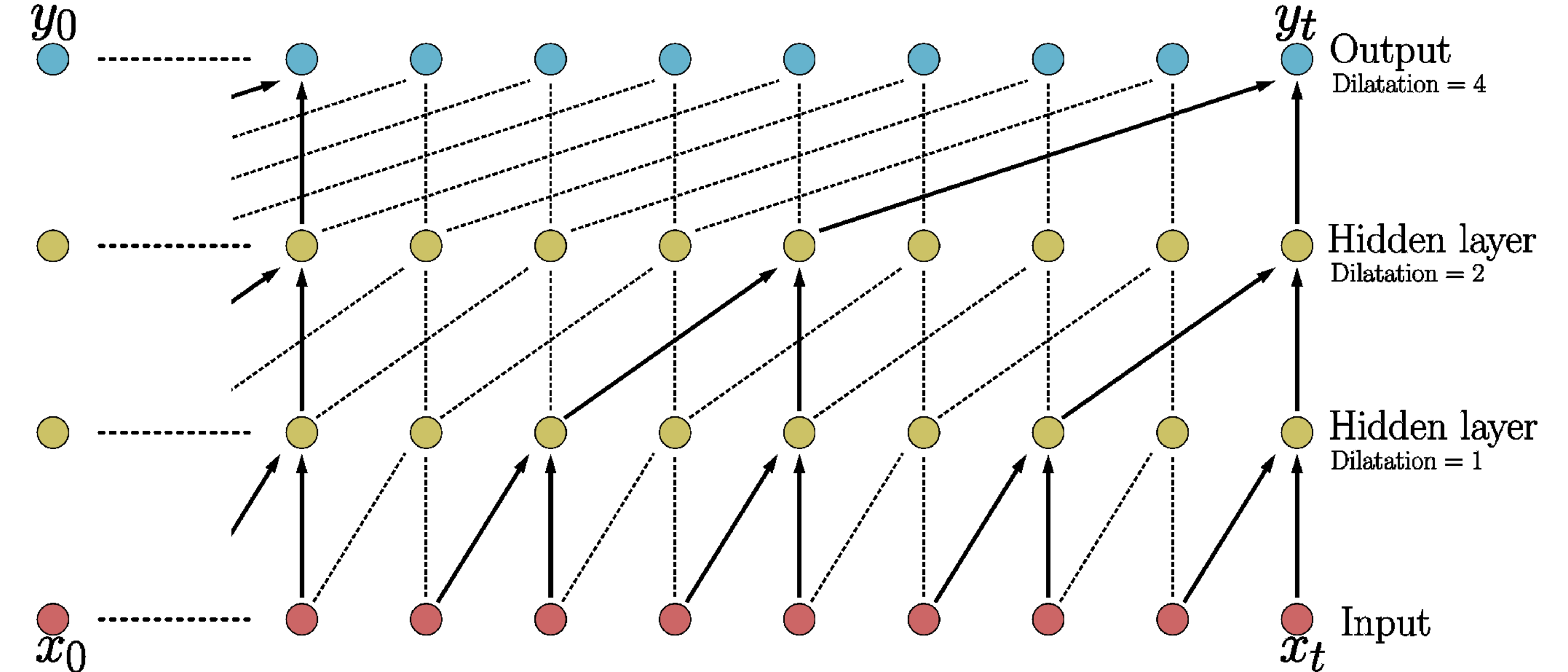}    
  \end{center}  
  \caption{Schematic representation of a Temporal Convolutional Network (TCN).
}
  \label{TCN}
\end{figure} 

The third architecture we use consists in TCN networks \cite{TCN}, which  use causal convolutions, meaning that at time step $t$, the network can only access data points at time step $t$ and earlier, ensuring no information leakage from future to past (see Fig. \ref{TCN}). The ability of causal convolution to keep previous data points in memory depends linearly on the depth of the network. This is why we are using dilated convolution to model long term dependencies of our system as shown in \cite{Oord} as it enables an exponentially large receptive field depending on the depth of the network. This enables TCN to function in a way similar to RNN.  For an input sequence $\textbf{U}$ of size $T$ (with elements $u_n$),  the dilated causal convolution $H$ we use is defined as
\begin{equation}
H(u)_n = (U*_dh)(u)_n = \sum_{i=0}^{k-1} h(i)u_{n-d.i}
\end{equation}
where $d$ is the dilatation factor, $h$ is a filter $\in\mathbb{R}^{k-1}$ where $k$ is the filter size and the indices $n-d.i$ represents the direction of the past. Using larger dilatation factor enables an output at the top level to represent a wider range of inputs, thus effectively expanding the receptive field of the network.

The LSTM and TCN networks are more complex networks and more demanding in computation than ESN. We have set up these networks with the Tensorflow library.  For a trajectory made of $N_i$ time step iterations, we use 35 successive points of the trajectory to predict the next step. In this way, we build a predicting vector of dimension $N_i-35$. We use batches of 32 successive values of this vector to update the network parameters with the gradient back propagation algorithm (using the Adam optimizer with an exponential  learning rate decay). This process is performed over all the values of the predicting vector, and repeated 10 times (number of epochs equal to 10). One has indeed to make several passes on the training data to get good results. On average, an epoch takes 30 seconds. Testing the different possible architectures therefore takes more time than for the ESN.

}


\end{document}